\documentclass[conference]{IEEEtran}
\IEEEoverridecommandlockouts
\usepackage{cite}
\usepackage{amsmath,amssymb}
\usepackage{algorithmic}
\usepackage{graphicx}
\usepackage{textcomp}
\usepackage{xcolor}
\usepackage[most]{tcolorbox}
\usepackage{tikz}
\usepackage[dvipsnames]{xcolor}
\usepackage[hidelinks]{hyperref}
\usepackage{makecell}
\usepackage{tabularx}
\usepackage{array}
\usepackage{makecell}
\usepackage{afterpage}
\usepackage{booktabs}
\usepackage{hyperref}
\usepackage{lmodern}
\usepackage{times}     
\hypersetup{
    colorlinks=true,
    urlcolor=blue
}
\usepackage{caption}
\usepackage{makecell} 

\usepackage{bm}
\usepackage{amsthm}
\usepackage{tabularx}

\usepackage{booktabs}
\usepackage{makecell}
\usepackage[table]{xcolor}
\usepackage{threeparttable}

\theoremstyle{definition}
\newtheorem{definition}{Definition}

\theoremstyle{remark}

\def\BibTeX{{\rm B\kern-.05em{\sc i\kern-.025em b}\kern-.08em
    T\kern-.1667em\lower.7ex\hbox{E}\kern-.125emX}}
\begin{document}

\title{Remote ID Spoofing-Aware Trajectory Planning for Small Unmanned Aerial Systems\\

}

\author{
\IEEEauthorblockN{
Jeremiah  Webb\textsuperscript{1},
Bryce Bjorkman\textsuperscript{1},
Abel Diaz Gonzalez\textsuperscript{1},
Austin Coursey\textsuperscript{1}\\
Noah Dahle\textsuperscript{1},
Kailani Lemieux Mack\textsuperscript{1},
Filippos Fotiadis\textsuperscript{2},
Gautam Biswas\textsuperscript{1},
Bryan C. Ward\textsuperscript{1},
Abenezer Taye\textsuperscript{1}
}

\IEEEauthorblockA{\textsuperscript{1}Vanderbilt University, Nashville, TN, USA}
\IEEEauthorblockA{\textsuperscript{2}The University of Texas at Austin, Austin, TX, USA}
}

\maketitle 

\begin{abstract}
This work presents a decentralized, spoofing-aware trajectory planning framework for small unmanned aerial systems operating under Remote Identification (RID) location spoofing attacks. Existing planners typically assume RID broadcasts are trustworthy, which can increase the risk of loss of separation and mid-air collisions when spoofing occurs. In contrast, the proposed approach explicitly treats RID information as unverified and incorporates physical-layer observations to assess broadcast credibility. Received signal-strength measurements from neighboring aircraft are used to detect spoofing and probabilistically localize a spoofing agent. The resulting uncertainty is converted into a risk-bounded unsafe region using a chance-constrained formulation and integrated into a per-agent Markov decision process–based planner. This enables real-time, decentralized collision avoidance while preserving mission objectives and scalability. Simulation results in a multi-aircraft package delivery scenario demonstrate reduced near mid-air collision events compared to planners that assume truthful RID data, while maintaining computational efficiency suitable for real-time execution. \textcolor{red}{[Code]}\footnote{\url{https://github.com/brycethebjorkman/uli-net-sim/tree/spoofing-aware-trajectory-planning}}
\end{abstract}

\section{Introduction} \label{sec:intro}

Small Unmanned Aerial Systems (sUAS) are increasingly deployed in a wide range of civilian and commercial applications, including surveillance, precision agriculture, and package delivery. The safe and scalable integration of these aircraft into shared low-altitude airspace relies on the use of Remote Identification (RID), a Federal Aviation Administration-mandated broadcast-based communication protocol that transmits aircraft state information, such as position, velocity, and identification, using unencrypted and unauthenticated wireless messages. While RID improves situational awareness, accountability, and scalability of sUAS operations, its lack of built-in security mechanisms exposes these operations to significant cyber vulnerabilities \cite{sharifi2026survey}.

\begin{figure}
    \centering
    \includegraphics[width=0.95\linewidth]{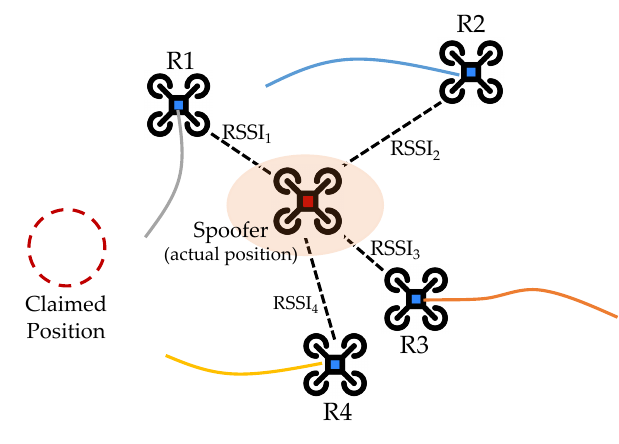}
    \caption{Spoofing-aware trajectory planning framework.}
    \label{fig:framework_schematic}
\end{figure}

Because RID is a relatively recent regulatory requirement, research specifically addressing RID spoofing is limited. However, a broader body of literature has investigated the problem of verifying the true location of an aircraft from signal observations to detect inconsistencies between claimed and physical positions. Among these,  \cite{sharifi2026survey} and \cite{blasch2019cyber} provide a comprehensive overview of RID spoofing attacks and defense strategies. However, the proposed countermeasures are mostly authentication-based, which violates the plain text broadcast assumption of RID. \cite{keizer2024ghostbuster} uses multilateration with synchronized receivers to estimate transmitter location and compare it with broadcast claims. However, the approach requires dense, time-synchronized ground receivers. \cite{sciancalepore2024orion} analyzes temporal consistency of broadcast messages to detect anomalous trajectories with a goal of validating reported trajectories over time, but their method assumes that broadcast data is mostly trustworthy. \cite{bjorkman2026remote} studies RID spoofing detection using Received Signal Strength Indicator (RSSI) inconsistencies and geometric constraints. Similarly, \cite{tian2025gnss} leveraged receiver motion-induced Doppler shifts to detect GNSS spoofing attacks, while \cite{shafique2021detecting} proposed data-driven methods for spoofing detection. However, these studies focus only on detection and do not address the problem of dynamically estimating the true location of the spoofer. In this work, we adopted a state estimator-based approach, combined with range-based multilateration, to estimate the true location of the spoofer aircraft, as illustrated in Fig. \ref{fig:framework_schematic}.  

On the other hand, there is a rich body of literature for multi-agent trajectory planning methods. Decentralized methods, where each aircraft resolves conflicts locally, are generally preferred over centralized planners due to their robustness and scalability. Among these, \cite{wu2022risk} formulates a chance-constrained trajectory planning problem that converts uncertainty into risk-bounded unsafe regions and enforces probabilistic safety guarantees. \cite{taye2024safe} proposes an MDP-based trajectory planning framework that enables scalable real-time decision-making by balancing safety and goal progress. \cite{fiorini1998motion} introduces the velocity obstacle framework for collision avoidance in dynamic environments by predicting future conflicts based on relative motion.  \cite{van2011reciprocal} extends this idea to reciprocal collision avoidance, where agents cooperatively adjust their velocities to avoid collisions.  \cite{sharon2015conflict} proposes conflict-based search for multi-agent trajectory planning, which resolves collisions through hierarchical constraint refinement. However, it is primarily designed for discrete environments and assumes full knowledge of agent states. Similarly, \cite{vinod2024decentralized} presented a reinforcement learning approach combined with robust optimization to achieve probabilistic safety guarantees under uncertainty. Further, \cite{lindqvist2021scalable} has shown that a distributed model predictive control approach can achieve high scalability in handling multi-agent collision avoidance. All of these trajectory planners rely on accurate knowledge of neighboring agent states and do not consider adversarially falsified RID broadcasts.

The key contribution of this work is enabling safe trajectory planning in the context where RID broadcasts are not inherently trustworthy due to location spoofing attacks. We proposed a framework that estimates the true location of an aircraft using physical-layer observations and generates safe trajectories for all aircraft in the environment. The proposed framework consists of three tightly coupled components: $(i)$ spoofing detection, $(ii)$ spoofer localization and unsafe area determination under uncertainty, and $(iii)$ risk-aware trajectory planning. The aircraft are only expected to coordinate in spoofing detection and localization with a ground control station, which computes spoofer state estimates and broadcasts the unsafe regions to avoid, while agents maintain independent trajectory planning and execution, eliminating the need for a centralized planner. 



The remainder of this paper is organized as follows.  Section \ref{sec:background} provides a brief discussion of the necessary background material related to RID broadcasts. Section \ref{sec:formulation} presents the problem formulation and provides an overview of our approach. In Section \ref{sec:spoofing}, we describe how the spoofing detection and localization scheme works. We then discuss how the risk-domain region is constructed and how the safe trajectories are computed in Section \ref{sec:trajectoryPlanning}. Section \ref{sec:methodology} explains the simulation architecture. Finally, Section \ref{sec:resDis} presents the implemented drone package delivery scenario and evaluates safety and computational runtime performance. The paper concludes with Section  \ref{sec:conclusion}.

\section{Background} \label{sec:background}
In this section, we outline the background material related to RID broadcasts, which are the central focus of this study.

\subsection{Remote ID}

The Federal Aviation Administration (FAA) requires unmanned aircraft operating in U.S. airspace to periodically broadcast Remote Identification (RID) messages. These broadcasts provide situational awareness information to nearby observers and airspace management systems. A standard Remote ID–enabled aircraft must transmit the following message elements \cite{faaRID}:

\begin{tcolorbox}
\begin{enumerate}
    \item Aircraft identity, consisting of either:
    \begin{enumerate}
        \item A manufacturer-assigned serial number, or
        \item A session ID.
    \end{enumerate}
    \item Latitude and longitude of the control station.
    \item Geometric altitude of the control station.
    \item Latitude and longitude of the unmanned aircraft.
    \item Geometric altitude of the unmanned aircraft.
    \item Velocity of the unmanned aircraft.
    \item A time mark corresponding to the Coordinated Universal Time (UTC) of the position measurement.
    \item Emergency status of the unmanned aircraft.
\end{enumerate}
\end{tcolorbox}

Many small unmanned aircraft systems (sUAS) lack onboard sensing capable of reliably detecting nearby aircraft. Consequently, RID information can be used to support collision avoidance and situational awareness. However, RID is susceptible to spoofing attacks in which a malicious transmitter broadcasts false state information. Such attacks can lead to loss of separation and reduced operational safety, potentially compromising mission integrity.





\section{Problem Formulation} \label{sec:formulation}

\subsection{Problem Description}

We consider an unstructured urban airspace in which a decentralized, homogeneous fleet of multirotor aircraft operates in a free-flight manner across a city-scale environment. Each aircraft is tasked with delivering a package from an initial warehouse location to a designated delivery site. In addition, we model a spoofing aircraft with a trajectory that is not known a priori and that broadcasts a falsified RID position. The objective of this study is to estimate the true location of this aircraft and generate safe trajectories for the benign aircraft in the environment. These trajectories enable the benign aircraft to avoid collision with the spoofer aircraft and reach their destinations safely. 

\subsection{Aircraft Model}\label{aircraft_model}

Each aircraft evolves in discrete time according to
\begin{equation}\label{eqn:generic_model}
\bm{\zeta}_{t+1} = f(\bm{\zeta}_t,\bm{u}_t),
\end{equation}
where $\bm{\zeta}_t \in \mathcal{X} \subset \mathbb{R}^n $ denotes the aircraft state at time step $t \in \mathbb{Z}^+$, and $\bm{u}_t \in \mathcal{U} \subset \mathbb{R}^m $ is the control input at time step t, with $\mathcal{U}$ being the action set containing all the control action combinations for an aircraft. The state $\bm{\zeta}_t$ consists of position, velocity, attitude, and angular rates. The control input is given by the total thrust and body torques, $\bm{u}_t = (T_t, \tau_{\phi_t}, \tau_{\theta_t}, \tau_{\psi_t})$. The function $f$ is obtained by discretizing a continuous-time six-degrees-of-freedom (6-DoF) dynamic model of an octorotor \cite{corbetta2019real}. The underlying continuous-time dynamics is given by

\begin{equation}
\dot{\bm{\zeta}} =
\left\{
\begin{aligned}
\ddot{x} &= (\text{s}\theta\,\text{c}\psi\,\text{c}\phi + \text{s}\phi\,\text{s}\psi)\frac{T}{m_t}, \\
\ddot{y} &= (\text{s}\theta\,\text{s}\psi\,\text{c}\phi - \text{s}\phi\,\text{c}\psi)\frac{T}{m_t}, \\
\ddot{z} &= -g + \text{c}\phi\,\text{c}\theta\,\frac{T}{m_t}, \\
\dot{\phi} &= p + q\,\text{s}\phi\,\text{t}\theta + r\,\text{c}\phi\,\text{t}\theta, \\
\dot{\theta} &= q\,\text{c}\phi - r\,\text{s}\phi, \\
\dot{\psi} &= q\,\frac{\text{s}\phi}{\text{c}\theta} + r\,\frac{\text{c}\phi}{\text{c}\theta}, \\
\dot{p} &= \frac{I_{yy}-I_{zz}}{I_{xx}}\,qr + \frac{l}{I_{xx}}\tau_\phi, \\
\dot{q} &= \frac{I_{zz}-I_{xx}}{I_{yy}}\,pr + \frac{l}{I_{yy}}\tau_\theta, \\
\dot{r} &= \frac{I_{xx}-I_{yy}}{I_{zz}}\,pq + \frac{l}{I_{zz}}\tau_\psi,
\end{aligned}
\right.
\end{equation}
where $\text{s}$, $\text{c}$, and $\text{t}$ represent $\sin$, $\cos$, and $\tan$ functions, respectively. The thrust $T$ is selected from nine logarithmically spaced values between $50$ and $500$ N, and each torque component is selected from seven uniformly spaced values between $-150$ and $150$ Nm.

\textbf{Assumption 1 (Communication and sensing model).}
All agents are RID-enabled and periodically broadcast state information at a fixed transmit power. The aircraft ID information in the RID messages are assumed to be authentic.

\textbf{Assumption 2 (Aircraft behavior and system model).}
All agents are cooperative and share identical multirotor dynamics, except for a single adversarial aircraft performing a spoofing attack, whose dynamics remain unknown to the cooperative agents. Unless spoofing is detected, RID-reported states are treated as reliable by neighboring agents.


\section{Spoofing Detection and Localization}
\label{sec:spoofing}

\subsection{Spoofer Detection}
\label{subsec:spooferDetection}

Because RID messages are not inherently trustworthy, additional observations are often used to assess their consistency. In this paper, we use physical-layer measurements, such as the received signal strength indicator (RSSI), to provide an independent source of information about the transmitter’s location. These measurements can be used to infer an estimate of the transmitter position, which is then compared with the position reported in the RID message. The discrepancy between the inferred and reported states serves as a way to verify the truthfulness of the broadcast message. We refer to this process as RID spoofing detection.

In this work, we adopt a spoofing detection scheme from \cite{bjorkman2026remote}. Here, we provide a brief summary of how the approach works. The method uses a Kalman filter to monitor the consistency between predicted and observed measurements through the innovation sequence. If a significant deviation, quantified by the Normalized Innovation Squared (NIS), is observed, then there is a mismatch between the physical-layer observation and the expected location of the transmitter. Thus, spoofing is detected.

\subsection{Spoofer Localization}
\label{subsec:spooferLocalization}

Once spoofing is detected, a key step in the safe trajectory planning process is dynamically estimating the true location of the malicious aircraft. We refer to this process as spoofer localization. We formulate the spoofer localization problem as a state estimation problem, where an interacting multiple model uses different motion hypotheses for the spoofer in the prediction step, and the position of the spoofer computed from a range-based multilateration approach is used in the correction step. In this subsection, we discuss how each component is formulated and implemented.

\subsubsection{Interacting  Multiple Model}
A key challenge in estimating the true location of the spoofer is the lack of knowledge of the spoofer aircraft dynamics and its intended maneuver. To address this challenge, we adopt a Bayesian filtering method known as the IMM filter \cite{barshalom2001tracking}. The intuition behind the IMM filter is to use multiple filters in parallel to track the target, each with a different motion hypothesis. The final estimate is computed by assigning different weights to each estimate.

In this study, we adopt constant velocity (CV) and constant acceleration (CA) motion hypotheses for the spoofer. The system and measurement models of these hypotheses have a general structure of the form

\begin{equation}\label{IMM_equations}
\left\{
\begin{aligned}
\mathbf{x}_{t+1}^{(m)} &= \mathbf{F}^{(m)} \mathbf{x}_t^{(m)} + \mathbf{w}_t^{(m)}, 
& \mathbf{w}_t^{(m)} &\sim \mathcal{N}(\mathbf{0}, \mathbf{Q}^{(m)}), \\
\mathbf{z}_t &= \mathbf{H}^{(m)} \mathbf{x}_t^{(m)} + \mathbf{v}_t, 
& \mathbf{v}_t &\sim \mathcal{N}(\mathbf{0}, \mathbf{R}).
\end{aligned}
\right.
\end{equation}
where $m \in \{\text{CV, CA} \}$,  $\mathbf{F}$ and $\mathbf{H}$ are the system and measurement models, and $ \mathbf{w}$ and $\mathbf{v}$ denote the process and measurement noises, with $\mathbf{Q}$ and $\mathbf{R}$ being the corresponding process and measurement noise covariances.

We now describe how the system and measurement models for the general structure described in \eqref{IMM_equations} are defined for both CV and CA motion hypotheses.

\paragraph{\underline{Constant Velocity (CV) Model}}
The equation of motion for the CV model is defined as a linear prediction of the spoofer’s state from a previous time step to the time of the measurements, where the state vector is composed of the linear position and velocity components. The system and measurement models for the CV model are given as:
\begin{equation}
\mathbf{F}^{\mathrm{CV}} =
\begin{bmatrix}
\mathbf{I}_3 & \Delta t\,\mathbf{I}_3 \\
\mathbf{0} & \mathbf{I}_3
\end{bmatrix}, 
\quad
\mathbf{H}^{\mathrm{CV}} = [\mathbf{I}_3 \;\; \mathbf{0}_{3\times3}].
\end{equation}

\paragraph{\underline{Constant Acceleration (CA) Model}}
The equation of motion for the CA model is defined as a linear prediction of the spoofer’s state from a previous time step to the time of the measurements, where the state vector is composed of the linear position, velocity, and acceleration components. The system and measurement models for the CA model are given~as:
\begin{equation}
\mathbf{F}^{\mathrm{CA}} =
\begin{bmatrix}
\mathbf{I}_3 & \Delta t\,\mathbf{I}_3 & \tfrac{1}{2}\Delta t^2\,\mathbf{I}_3 \\
\mathbf{0} & \mathbf{I}_3 & \Delta t\,\mathbf{I}_3 \\
\mathbf{0} & \mathbf{0} & \mathbf{I}_3
\end{bmatrix}, 
\quad
\mathbf{H}^{\mathrm{CA}} = [\mathbf{I}_3 \;\; \mathbf{0}_{3\times6}].
\end{equation}

At each time step, the two filters (i.e., those based on CV and CA motion hypotheses) execute the standard Kalman filter predict--update cycle and provide the state estimate $\hat{\mathbf{x}}_t^{(m)}$, covariance $\mathbf{P}_t^{(m)}$, and model probability $\mu_t^{(m)}$ (computed from the innovation likelihood). The fused estimate is then obtained by combining the outputs of these two filters as
\begin{equation}
\hat{\mathbf{x}}_t = \sum_{m} \mu_t^{(m)} \hat{\mathbf{x}}_t^{(m)},
\end{equation}

\begin{equation}
\mathbf{P}_t =
\sum_{m} \mu_t^{(m)} \Bigl(
\mathbf{P}_t^{(m)} +
(\hat{\mathbf{x}}_t^{(m)} - \hat{\mathbf{x}}_t)
(\hat{\mathbf{x}}_t^{(m)} - \hat{\mathbf{x}}_t)^\top
\Bigr).
\label{eq:imm_merge_rewrite}
\end{equation}
These fused position estimate and covariance values are used as the spoofer position estimate for downstream planning.

A key parameter in the IMM filter that defines the probability that the spoofer switches from one motion hypothesis to another is the state transition matrix. The specific state transition matrix and all other IMM filter related parameters used in this paper are presented in Table \ref{tab:imm_parameters_appendix}.


\subsubsection{Multilateration} The measurement used for the update stage of the IMM filter ($\mathbf{z}_t$) is obtained via a process known as federated multilateration. In this approach, the position of the spoofer is determined when at least four receivers are available (three positional unknowns and transmission power). The multilateration procedure depends on computing the range estimates inferred from a free-space path loss model of wireless signal propagation  \cite{rappaport2002wireless}:
\begin{equation}
P_{r}(d) = P_{t} - 20\log_{10}(d) - 20 \log_{10}(\hat{f}) - 32.44
\label{eq:fspl_rewrite}
\end{equation}
where $d$ denotes the transmitter-receiver distance and $\hat{f}$ denotes transmission frequency. In addition, $P_{r}$ and $P_{t}$ denote received and transmission powers, respectively. In this study, we assume the RID messages are transmitted at $2.4$ GHz. When at least four receivers observe a transmission, the GCS performs multilateration by jointly estimating transmitter position in 3D space ($\bm{x} \in \mathbb{R}^3$) and transmit power $P_{\mathrm{t}}$ via nonlinear least squares,
\begin{equation}
\min_{\bm{x},\, P_{\mathrm{t}}}
\sum_{i}
\left(
{P_r}_i - \left(
P_{\mathrm{t}} - 20\log_{10}\|\bm{x} - \bm{r}_i\| - 40.04
\right)
\right)^2,
\label{eq:mlat_detect_rewrite}
\end{equation}
where $\bm{r}_i$ are the known positions of the receiving agents. The above optimization problem is solved using a non-linear least-squares solver based on the Trust Region Reflective (TRF) algorithm, as implemented in SciPy \cite{scipy_least_squares}. Because the cost function is non-convex due to the nonlinear dependence of the received power on the logarithm of the Euclidean distance $\|\bm{x} - \bm{r}_i\|$, the optimization may admit multiple local minima, and solution quality depends on initialization. Consequently, we initialized the solver from the current estimate or the centroid of receiver positions.

\section{Risk-Aware Trajectory Planning} \label{sec:trajectoryPlanning}
In this section, we discuss the last component of the proposed framework that generates the safe trajectories. 

\subsection{Chance-Constrained Unsafe Region}\label{CC}

After obtaining a probabilistic estimate of the spoofer location, we convert it into a deterministic unsafe region using a chance-constrained formulation. The following definition adopts the formulation in \cite{wu2022risk}.

\begin{definition}[Chance-constrained unsafe region]
Let $\hat{\mathbf{x}}_{t} \in \mathbb{R}^3$ and $\mathbf{P}_{t} \in \mathbb{R}^{3\times3}$ denote the mean and covariance of the spoofer location estimate obtained from the IMM filter at time step $t$. Let $\mathbf{x}_{t} \in \mathbb{R}^3$ denote the position of the aircraft.

For a given risk level $\alpha \in (0,1)$, the chance-constrained safe set is defined as
\begin{equation}
(\mathbf{x}_{t} - \hat{\mathbf{x}}_{t})^\top
\mathbf{P}_{t}^{-1}
(\mathbf{x}_{t} - \hat{\mathbf{x}}_{t})
>
F^{-1}_{\chi^2_3}(1 - \alpha),
\label{eq:chance_constraint}
\end{equation}
and the corresponding unsafe region is given by the complement of this set.
\end{definition}
The threshold $F^{-1}_{\chi^2_3}(1-\alpha)$ denotes the inverse cumulative distribution function of a chi-squared random variable with three degrees of freedom, corresponding to the three-dimensional uncertainty in the spoofer location estimate.
Satisfaction of \eqref{eq:chance_constraint} ensures that the probability of the aircraft entering the unsafe region does not exceed $\alpha$, i.e.,
\begin{equation}
\Pr\!\left(
(\mathbf{x}_t - \hat{\mathbf{x}}_{t})^\top
\mathbf{P}_{t}^{-1}
(\mathbf{x}_t - \hat{\mathbf{x}}_{t})
\leq
F^{-1}_{\chi^2_3}(1 - \alpha)
\right)
\leq \alpha,
\end{equation}
which implies that $\mathbf{x}_t$ lies outside the risk domain with confidence $1-\alpha$.

Geometrically, the inequality defines an ellipsoidal unsafe region centered at $\hat{\mathbf{x}}_{t}$, whose size is determined by the spoofer location uncertainty $\mathbf{P}_{t}$ and the risk parameter $\alpha$. A well-conditioned covariance matrix, corresponding to favorable observer geometry, results in a compact and nearly isotropic ellipsoid. In contrast, poor geometries (e.g., near-coplanar observer configurations) lead to ill-conditioned covariance matrices, producing elongated ellipsoids that reflect increased uncertainty along certain directions.
\begin{figure*}[t]
\centering
\includegraphics[width=\textwidth]{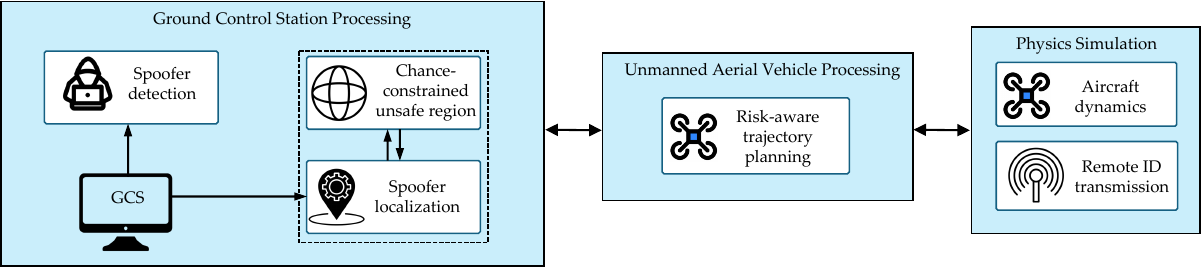}
\caption{System architecture. }
\label{fig:architecture}
\end{figure*}
\subsection{Trajectory Planner}\label{AA}

The trajectory planner implemented in this paper, introduced in \cite{taye2024safe}, formulates the decision-making problem as a Markov decision process (MDP). A significant update to the current planner, compared to \cite{taye2024safe}, is the use of chance-constrained formulation presented in Section \ref{CC} for collision avoidance with the true location of the spoofer aircraft. 

The state transition function and the action set used in the MDP formulation are the aircraft model and the control actions presented in Section \ref{aircraft_model}. Therefore, in this section, we focus on how the reward function is designed. 
\subsubsection{Reward Function} The reward function is the primary mechanism for controlling the behavior of an MDP agent. In this work, we utilize both positive and negative rewards, as presented in Table \ref{tab:mdp_reward}, to guide the aircraft to their destination while avoiding potential collisions with the spoofer aircraft and other nearby benign aircraft. As shown in Table \ref{tab:mdp_reward}, the positive rewards are placed at the assigned destinations, and the negative rewards are assigned within the chance-constrained unsafe region of the spoofer aircraft. In addition, to prevent benign-to-benign aircraft collision, we placed a negative reward in the vicinity of nearby benign aircraft whenever the safe inter-aircraft distance threshold $(d_{ij} \leq d_{\text{safe}} )$ is violated.

\begin{table}
\centering
\small
\renewcommand{\arraystretch}{1.15}
\setlength{\tabcolsep}{3pt}
\caption{\textbf{Reward Function for Each Aircraft}}
\label{tab:mdp_reward}

\begin{tabularx}{\columnwidth}{l c X c}
\toprule
\textbf{Component} & \textbf{Reward} & \textbf{Region} & \textbf{Decay} \\
\midrule

Goal attraction 
& $200$ 
& Goal state 
& $0.999$ \\

Collision (benign) 
& $-(100t + 500)$ 
& $d_{ij} \leq d_{\text{safe}}$
& $0.97$ \\

Collision (spoofer) 
& $-1000$ 
& $\mathbf{x}_t \in \mathcal{X}_{\text{unsafe}}(\alpha)$ 
& $0.97$ \\

\bottomrule
\end{tabularx}
\end{table}

At each planning update, an aircraft generates a set of possible future states using its dynamic model and action set in a receding horizon manner. It then computes the value of each projected future state using the positive and negative rewards defined in Section \ref{tab:mdp_reward}. Finally, it selects the control action from the action set that maximizes the total reward.

\section{System Architecture} \label{sec:methodology}

The proposed spoofing-aware trajectory planning framework is implemented in the OMNeT++ simulator. OMNeT++ is a discrete-event simulator~\cite{omnet} with the INET Framework~\cite{inet} that simulates physical-layer signal propagation while accounting for factors that affect signal propagation, including distance, transmission power, and environmental conditions. However, multipath effects, channel throughput, and packet loss are not modeled in this work and are left for future investigation.

To facilitate the simulation, all parameters relevant to signal propagation, including the path loss model and the aircraft model, are implemented in OMNeT++ using the C++ programming language. Meanwhile, the risk-aware MDP-based trajectory planner is implemented in Python and re-plans the trajectories of each aircraft every $0.5$~s. To avoid coupling downstream behavior to variable detector trigger times, we use a forced spoofing alert for the designated serial number of the adversarial aircraft at $5$~s. As discussed in \cite{bjorkman2026remote}, open-loop detector timing depends strongly on scenario geometry; fixing the alert time isolates subsequent localization and planning effects for the experiments reported here.

Each aircraft transmits RID beacons at $1$~Hz, and RSSI observations are forwarded to the GCS for spoofing detection and localization. The GCS executes periodic updates every $0.25$~s and propagates a per-spoofer IMM state estimate at each decision-making step. The chance-constrained formulation is then used to convert the probabilistic estimate of the spoofer location into a deterministic unsafe region. The resulting unsafe-region parameters are broadcast to the aircraft and incorporated into onboard planning through reward functions designed to avoid unsafe areas. A schematic diagram depicting the system architecture is shown in Fig.~\ref{fig:architecture}.

\begin{figure*}[t!]
    \centering
    \includegraphics[width=\textwidth,height=0.85\textheight,keepaspectratio]{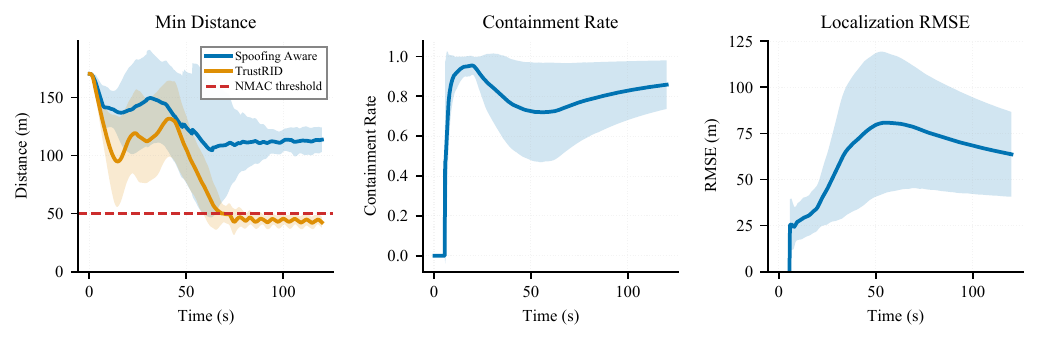}
    \caption{Performance metrics over time for the scenario with 16 agents. (Left) minimum distance to the spoofer; (Middle) containment rate; (Right) localization accuracy. Solid lines represent the mean values, while shaded regions indicate the range between the minimum and maximum values.}
    \label{fig:performance_overtime}
\end{figure*}

\section{Results and Discussion} \label{sec:resDis}
\subsection{Scenario Description}

We evaluate the proposed spoofing-aware trajectory planning framework in a drone package-delivery operation with multiple benign drones and one active spoofer. The operation is implemented within a $1$ km $\times$ $1$ km urban environment modeled after the FAA UTM framework for low-altitude airspace integration~\cite{faaUTM}. In this setting, there are three depot locations, which serve as hubs where drones are stationed for charging, maintenance, and package loading. A single scenario in this operation includes a varying number of traffic densities for the benign aircraft (4, 8, 12, 16), starting at the hubs and crossing the airspace to delivery sites. In addition, in each scenario, there is a single spoofer aircraft whose dynamic model and intended trajectory are unknown. In this scenario, the spoofer follows a straight-line trajectory to its goal while broadcasting a constant position offset from its true state in its RID messages and performing no collision avoidance.

To better evaluate the efficacy of the proposed framework, we use a trajectory planner that assumes all RID broadcast messages are truthful as a baseline. For each scenario, we run 30 experiments with matched seeds for the two planners\footnote{A sample video of the simulation for 8 aircraft can be found here: https://youtu.be/LTTAv5DEp5U}. These 30 runs enable us to statistically isolate the effects of randomness arising from communication (e.g., communication timing and radio frequency (RF) signal noise) and scenario-specific variabilities (e.g., random assignments of delivery destinations).

\subsection{Metrics}

The goal of this study is to develop a multi-agent trajectory planning framework that enables safe movement of aircraft in the presence of location spoofing attacks. To that end, the framework is expected $(i)$ to effectively identify the true location of the spoofer and track it throughout the entire duration of the operation, $(ii)$ to perform risk-aware trajectory planning for each aircraft with a user-defined risk threshold, and $(iii)$ to achieve this in a scalable manner. As such, we use the following metrics to evaluate the proposed framework and compare its performance with a baseline trajectory planner:

\begin{itemize}
\item \textbf{Localization containment rate (LCR):} indicates the fraction of time the true spoofer position is contained within the estimated chance-constrained ellipsoid. This is the primary performance indicator for the spoofer localization component of the framework.
\item \textbf{Near mid-air collisions (NMACs):} indicates events where separation is below $50$ m. This threshold is chosen based on a maximum aircraft speed of 8 m/s and a minimum separation time of 3 seconds for two aircraft heading towards each other. We report three variants of NMAC: $(i)$ benign--benign pairwise proximity, $(ii)$ benign--true-spoofer proximity, and $(iii)$ benign entry into the spoofing unsafe (chance-constrained) ellipsoid.
\item \textbf{Computational cost:} the time taken by the framework to complete a flight mission for all aircraft involved.
\end{itemize}

\subsection{Results and Discussion}\label{sec:results}

\subsubsection{Safety}
We evaluated our proposed framework across different traffic levels, and in the presence of a single spoofer. The NMAC results are provided in Table~\ref{tab:merged_clean}. From the results, it can be seen that our approach showed substantial NMAC performance improvement compared to the baseline. Specifically, the total NMAC means were lower under our approach for every scenario. In addition, the strongest safety improvement appears in the benign--spoofer interaction metric. In the moderate- and high-density cases, our proposed approach substantially suppresses close encounters with the malicious aircraft.

This traffic density-dependent widening of the performance gap observed between our approach and the baseline indicates that risk-aware avoidance is most beneficial precisely when interaction pressure is highest. The 4-agent case remains a low-event regime (near-zero benign--spoofer NMACs in both planners due to sparse geometry), suggesting that the principal advantage of our approach emerges as traffic and encounter opportunities increase.

\subsubsection{Localization Containment}
\label{sec:localization_containment}

Localization quality, measured by the spoofer containment rate, improved with observer density, as presented in Table~\ref{tab:merged_clean}. This trend is consistent with improved RSSI observer geometry as more benign agents contribute measurements. This phenomenon is also supported by the accuracy of the localization scheme, assessed through root mean squared error (RMSE) and mean absolute error (MAE) values with respect to the ground truth. These results indicate that position error improves with density and that the largest geometric and observability limitations occur in the minimum-observer case. The results are consistent with the previous observation, i.e., the more the number of aircraft in the environment, the better the localization accuracy.

Furthermore, the maximum unsafe-ellipsoid radius remains relatively stable across scenarios, with a modest increase in size as the number of aircraft increases. Together, these results suggest that while four observers satisfy the minimum requirement for RSSI NLLS multilateration, this configuration is underconstrained in practice and yields high-variance localization; hence, adding observers substantially improves both localization accuracy and containment reliability.

\begin{table*}[t]
\centering
\footnotesize
\renewcommand{\arraystretch}{1.15}
\setlength{\tabcolsep}{5pt}
\caption{\textbf{Safety and spoofer-localization metrics across different traffic densities}}
\label{tab:merged_clean}

\begin{tabular}{lcccc}
\toprule
\textbf{Metric} & \textbf{4 Agents} & \textbf{8 Agents} & \textbf{12 Agents} & \textbf{16 Agents} \\
\midrule

\rowcolor{gray!15}
\multicolumn{5}{l}{\textbf{Safety Metrics (SA)}} \\

Total NMACs
& $0.300 \pm 0.466$ & $1.733 \pm 1.230$ & $4.067 \pm 2.545$ & $7.000 \pm 3.311$ \\
Benign--Benign
& $0.267 \pm 0.450$ & $1.633 \pm 1.273$ & $3.700 \pm 2.409$ & $6.767 \pm 3.245$ \\
\textbf{Benign--Spoofer}
& \bm{$0.033 \pm 0.183$} & \bm{$0.100 \pm 0.305$} & \bm{$0.367 \pm 0.556$} & \bm{$0.233 \pm 0.568$} \\
Unsafe violations
& $2.700 \pm 3.669$ & $0.933 \pm 1.230$ & $1.533 \pm 2.177$ & $3.033 \pm 2.965$ \\

\midrule
\rowcolor{gray!10}
\multicolumn{5}{l}{\textbf{Safety Metrics (RID)}} \\

Total NMACs
& $0.267 \pm 0.450$ & $1.933 \pm 1.507$ & $4.933 \pm 1.929$ & $8.967 \pm 3.264$ \\
Benign--Benign
& $0.267 \pm 0.450$ & $1.200 \pm 1.064$ & $3.533 \pm 1.655$ & $6.767 \pm 2.967$ \\
\textbf{Benign--Spoofer}
& \bm{$0.000 \pm 0.000$} & \bm{$0.733 \pm 0.691$} & \bm{$1.400 \pm 0.814$} & \bm{$2.200 \pm 0.805$} \\
Unsafe violations
& -- & -- & -- & -- \\

\midrule
\midrule
\rowcolor{gray!15}
\multicolumn{5}{l}{\textbf{Localization Metrics (SA only)}} \\

\textbf{Containment rate}
& \bm{$0.478 \pm 0.259$} & \bm{$0.804 \pm 0.186$} & \bm{$0.894 \pm 0.109$} & \bm{$0.839 \pm 0.106$} \\
RMSE (m)
& $334.394 \pm 434.679$ & $71.953 \pm 32.075$ & $57.274 \pm 22.065$ & $63.649 \pm 15.694$ \\
MAE (m)
& $239.724 \pm 308.027$ & $61.286 \pm 26.111$ & $48.596 \pm 16.370$ & $51.190 \pm 11.357$ \\
Ellipsoid radius (m)
& $81.687 \pm 6.848$ & $87.534 \pm 2.338$ & $87.334 \pm 2.075$ & $87.885 \pm 1.184$ \\

\bottomrule
\end{tabular}
\end{table*}

Finally, in Fig. \ref{fig:performance_overtime}, we report the progression of these safety and spoofer localization performance metrics over time for 16 aircraft to better illustrate the performance of the proposed framework. As shown in the figure, the minimum distance to the spoofer remains above the NMAC threshold for the proposed framework. In contrast, the baseline trajectory planner incurs significant safety violations once the environment becomes crowded. In addition, the containment rate and the localization performances indicate that the proposed framework was able to consistently track the true location of the spoofer over time. 

\begin{table}[!t]
\centering
\small
\renewcommand{\arraystretch}{1.15}
\setlength{\tabcolsep}{3pt}
\caption{\textbf{Computation time (sec)}}
\label{tab:runtime_by_scenario_sa_rid}

\begin{tabular}{lcccc}
\toprule
\rowcolor{gray!15}
& \multicolumn{2}{c}{\textbf{Total wall-clock}} &
  \multicolumn{2}{c}{\textbf{GCS compute}} \\
\cmidrule(lr){2-3} \cmidrule(lr){4-5}
\rowcolor{gray!15}
\textbf{Agents} &
\textbf{SA} &
\textbf{Trust-RID} &
\textbf{SA} &
\textbf{Trust-RID} \\
\midrule
4  & $40.07 \pm 3.70$   & $31.53 \pm 4.61$   & $5.42 \pm 0.68$ & $0.14 \pm 0.02$ \\
8  & $83.60 \pm 21.94$  & $82.57 \pm 14.72$  & $4.66 \pm 0.33$ & $0.36 \pm 0.06$ \\
12 & $188.40 \pm 48.08$ & $153.70 \pm 27.35$ & $6.27 \pm 0.92$ & $0.58 \pm 0.12$ \\
16 & $221.77 \pm 39.36$ & $108.23 \pm 13.56$ & $7.24 \pm 1.11$ & $0.39 \pm 0.04$ \\
\bottomrule
\end{tabular}

\vspace{2pt}
\footnotesize
Total wall-clock is the full single-process simulation time over
$120$\,s of flight; GCS compute isolates the spoofing-aware detection,
localization, and chance-constrained processing.
\end{table}

\subsubsection{Computational Cost}
Simulations were conducted on a dual-socket Intel Xeon Silver 4314 platform (2$\times$16 cores, 64 logical CPUs at 2.4~GHz) with 64~GB of RAM. Runtime statistics are reported in Table~\ref{tab:runtime_by_scenario_sa_rid}. Computational cost was evaluated by measuring wall-clock runtime for each scenario while keeping the simulation horizon fixed at 120 simulated seconds. We additionally report the GCS compute time, which isolates the spoofing-aware processing (detection, localization, and chance-constrained unsafe-region construction) from the cost of the full simulation. This setup isolates how computational cost scales with problem size.

The results show that the proposed approach incurs additional overhead relative to the baseline due to the inclusion of spoofing detection, localization, and risk-aware decision-making within the planning loop. Nevertheless, the framework exhibits near-linear scaling of runtime with respect to the number of aircraft in the environment. Furthermore, although absolute runtime increases with traffic density for both planners, the relative overhead of the proposed approach decreases at higher agent counts. The GCS compute time remains small and nearly flat across densities (roughly 5--7~s over 120~s of flight), indicating that the spoofing-aware computation contributes little to the overall runtime.

Overall, the results indicate that the proposed spoofing-aware trajectory planning improves safety under RID spoofing while remaining practical for real-time use. Relative to the baseline trajectory planner that assumes all RID messages are truthful, our approach consistently lowers both total and spoofing-specific NMACs and increases separation from the malicious transmitter. Moreover, although four observers satisfy the minimum requirement of the localization pipeline, this case is often poorly conditioned geometrically, leading to large localization errors and impacting safety.

\subsection{Practical Considerations}

\subsubsection{Channel modeling fidelity}
The physical-layer simulation models distance-dependent attenuation
through the free-space path loss model in~\eqref{eq:fspl_rewrite}, but does not
yet capture multipath propagation, finite channel throughput, or packet
loss. In urban environments, reflections and shadowing introduce bias
and variance into RSSI measurements, which would
increase the range errors entering the multilateration
in~\eqref{eq:mlat_detect_rewrite} and degrade the localization accuracy reported in
Table~\ref{tab:merged_clean}. Packet loss can also reduce the number of
synchronized reports below the four-receiver minimum, worsening the
geometric conditioning already observed in the low-density cases. The
chance-constrained formulation mitigates this risk by inflating the
unsafe region in proportion to the estimate covariance, so degraded
measurements yield more conservative avoidance regions rather than
silent failures. Accounting for these effects in the propagation and estimation models
is therefore a natural next step toward deployment.

\subsubsection{Real-time feasibility}
While the total wall-clock runtimes in
Table~\ref{tab:runtime_by_scenario_sa_rid} exceed the $120$\,s
simulation horizon, they measure the cost of the full discrete-event
simulation, including OMNeT++/INET event processing, dynamics
integration, GCS estimation, and the planners for all agents, executed
sequentially within a single process. They therefore reflect simulation
cost rather than the per-component latency that governs real-time
execution. The spoofing-aware computation introduced in this work
resides in the GCS; in the 16-agent scenario it consumes only
$7.24 \pm 1.11$\,s over $120$\,s of flight, which, at the $0.25$\,s
update interval, corresponds to roughly $15$\,ms per update and is well
within the GCS cycle time. The remaining wall-clock time is dominated
by the serialized network and dynamics simulation and the per-agent
planners, which in deployment execute in parallel on separate onboard
computers rather than on a single host. The aggregate simulation
runtime therefore does not bound flight-time feasibility, and the
spoofing-aware additions operate comfortably in real time.

\subsubsection{Adaptive adversaries}
The spoofer in our evaluation follows a straight-line trajectory, but
the localization pipeline is not specialized to this case. The IMM
filter runs constant-velocity and constant-acceleration hypotheses with
a model transition matrix, so it is designed to track maneuvering,
nonlinear trajectories by reweighting and switching between motion
models as the target accelerates or turns; the straight-line spoofer is
thus a representative evaluation choice rather than an assumption
required by the method. Tracking a more aggressively maneuvering
adversary would, however, require re-tuning the IMM parameters in
Table~\ref{tab:imm_parameters_appendix}, such as the process-noise covariances and
the model transition probabilities, to match the expected maneuver
dynamics. Beyond its own motion, a strategic adversary
could attempt to degrade observer geometry, for example positioning
itself so that the contributing receivers become near-coplanar, which
ill-conditions the covariance $P_t$ and elongates the unsafe ellipsoid.
We expect the chance-constrained design to remain fail-safe under such
behavior, since poor geometry inflates $P_t$ and enlarges the avoided
region, consistent with the ellipsoid radius trend in
Table~\ref{tab:merged_clean}. 


\section{Conclusion} \label{sec:conclusion}
In this work, we present a spoofing-aware trajectory planning framework that enables safe movement of sUAS aircraft in the presence of RID spoofing attacks. The framework is comprised of three tightly interconnected components: spoofing detection, spoofing localization, and risk-aware trajectory planning. The results indicate that the proposed method achieves substantial safety performance improvements compared with a baseline trajectory planner that assumes all RID broadcast messages are truthful. In addition, although the proposed approach incurs additional computational overhead compared to the baseline trajectory planner, it remain scalable and able to support a high number of aircraft in the environment. Future works include incorporating optimal observer allocation and trajectory planning for efficient and accurate spoofer localization. Furthermore, additional physical layer properties, such as the Doppler shift and angle of arrival, will be incorporated for better spoofer localization accuracy.      

\section*{Acknowledgment}
This material is based upon work supported by the NASA Aeronautics Research Mission Directorate (ARMD) University Leadership Initiative (ULI) under cooperative agreement number 80NSSC24M0070. Any opinions, findings, and conclusions or recommendations expressed in this material are those of the author(s) and do not necessarily reflect the views of the National Aeronautics and Space Administration.

\bibliographystyle{IEEEtran}  
\bibliography{bibliography}

@inproceedings{sharifi2026survey,
  title={A Survey of Security Challenges and Solutions for UAS Traffic Management (UTM) and small Unmanned Aerial Systems (sUAS)},
  author={Sharifi, Iman and Ghazanfari, Mahyar and Taye, Abenezer and Wei, Peng and Ahmed, Maheed and Tae Kim, Hyeong and Ghasemi, Mahsa and Gupta, Vijay and Dahle, Noah W and Canady, Robert and others},
  booktitle={AIAA SCITECH 2026 Forum},
  pages={2892},
  year={2026}
}

@inproceedings{keizer2024ghostbuster,
  title={Ghostbuster: Detecting misbehaving remote id-enabled drones},
  author={Keizer, Mart and Sciancalepore, Savio and Oligeri, Gabriele},
  booktitle={2024 IEEE 21st Consumer Communications \& Networking Conference (CCNC)},
  pages={324--332},
  year={2024},
  organization={IEEE}
}

@article{sciancalepore2024orion,
  title={ORION: Verification of drone trajectories via remote identification messages},
  author={Sciancalepore, Savio and Davidovic, Filip and Oligeri, Gabriele},
  journal={Future Generation Computer Systems},
  volume={160},
  pages={869--878},
  year={2024},
  publisher={Elsevier}
}

@article{omnet,
    author      = {Varga, Andr�s},
    description = {Dissertation},
    journal     = {Proceedings of the European Simulation Multiconference (ESM'2001)},
    month       = {June},
    title       = {The OMNeT++ Discrete Event Simulation System},
    year        = {2001},
}

@misc{inet,
    author      = {},
    title       = {INET Framework},
    url         = {https://inet.omnetpp.org/},
    year        = {2024},
    note        = {[Accessed 01-05-2024]},
}

@article{vinod2024decentralized,
  title={Decentralized, Safe, Multiagent Motion Planning for Drones Under Uncertainty via Filtered Reinforcement Learning},
  author={Vinod, Abraham P. and Safaoui, Sleiman and Summers, Tyler H. and Yoshikawa, Nobuyuki and Di Cairano, Stefano},
  journal={IEEE Transactions on Control Systems Technology},
  year={2024},
  publisher={IEEE},
  doi={10.1109/TCST.2024.3123456}
}

@inproceedings{lindqvist2021scalable,
  title={A scalable distributed collision avoidance scheme for multi-agent UAV systems},
  author={Lindqvist, Bj{\"o}rn and Sopasakis, Pantelis and Nikolakopoulos, George},
  booktitle={2021 IEEE/RSJ International Conference on Intelligent Robots and Systems (IROS)},
  pages={9212--9218},
  year={2021},
  organization={IEEE},
  doi={10.1109/IROS51168.2021.9636293}
}

@inproceedings{bjorkman2026remote,
  title={Remote ID Spoofing Attacks and Defenses},
  author={Bjorkman, Bryce and Zheng, Stanley and Coursey, Austin and Lemieux-Mack, Cailani and Gonzalez, Samuel and Diaz-Gonzalez, Abel and Dahle, Noah W and Koroma, Neils and Canady, Robert E and Koutsoukos, Xenofon and others},
  booktitle={AIAA SCITECH 2026 Forum},
  pages={2665},
  year={2026}
}

@article{wu2022risk,
  title={Risk-bounded and Fairness-aware Path Planning for Urban Air Mobility under Uncertainty},
  author={Wu, Peng and Xie, Jun and Liu, Yifan and Chen, Jie},
  journal={Aerospace Science and Technology},
  volume={127},
  pages={107738},
  year={2022},
  doi={10.1016/j.ast.2022.107738}
}

@article{taye2024safe,
  title={Safe and scalable real-time trajectory planning framework for urban air mobility},
  author={Taye, Abenezer G and Valenti, Roberto and Rajhans, Akshay and Mavrommati, Anastasia and Mosterman, Pieter J and Wei, Peng},
  journal={Journal of Aerospace Information Systems},
  volume={21},
  number={8},
  pages={641--650},
  year={2024},
  publisher={American Institute of Aeronautics and Astronautics}
}

@inproceedings{fiorini1998motion,
  title={Motion Planning in Dynamic Environments Using Velocity Obstacles},
  author={Fiorini, Paolo and Shiller, Zvi},
  booktitle={IEEE International Conference on Robotics and Automation},
  pages={560--565},
  year={1998}
}

@article{van2011reciprocal,
  title={Reciprocal n-body Collision Avoidance},
  author={Van den Berg, Jur and Guy, Stephen J. and Lin, Ming and Manocha, Dinesh},
  journal={Robotics Research},
  pages={3--19},
  year={2011}
}

@inproceedings{sharon2015conflict,
  title={Conflict-Based Search for Optimal Multi-Agent Path Finding},
  author={Sharon, Guni and Stern, Roni and Felner, Ariel and Sturtevant, Nathan R.},
  booktitle={Artificial Intelligence},
  volume={219},
  pages={40--66},
  year={2015}
}

@inproceedings{corbetta2019real,
  title={Real-time UAV trajectory prediction for safety monitoring in low-altitude airspace},
  author={Corbetta, Matteo and Banerjee, Portia and Okolo, Wendy and Gorospe, George and Luchinsky, Dmitry G},
  booktitle={Aiaa aviation 2019 forum},
  pages={3514},
  year={2019}
}

@book{rappaport2002wireless,
  title={Wireless Communications: Principles and Practice},
  author={Rappaport, Theodore S.},
  edition={2nd},
  year={2002},
  publisher={Prentice Hall}
}

@misc{faaRID,
  author = {{Federal Aviation Administration}},
  title  = {Remote Identification of Unmanned Aircraft},
  howpublished = {Code of Federal Regulations, Title 14, Part 89},
  year   = {2021},
  url    = {https://www.ecfr.gov/current/title-14/chapter-I/subchapter-F/part-89}
}

@book{barshalom2001tracking,
  author    = {Bar-Shalom, Yaakov and Li, X. Rong and Kirubarajan, Thiagalingam},
  title     = {Estimation with Applications to Tracking and Navigation},
  publisher = {Wiley},
  year      = {2001}
}

@techreport{faaUTM,
  author       = {{Federal Aviation Administration}},
  title        = {Unmanned Aircraft System (UAS) Traffic Management (UTM) Concept of Operations v2.0},
  institution  = {Federal Aviation Administration},
  address      = {Washington, DC},
  month        = aug,
  year         = {2022},
  url          = {https://www.faa.gov/sites/faa.gov/files/2022-08/UTM_ConOps_v2.pdf}
}

@misc{scipy_least_squares,
  author       = {{SciPy Developers}},
  title        = {{scipy.optimize.least\_squares}: Nonlinear least-squares solver with Trust Region Reflective method},
  year         = {2024},
  howpublished = {\url{https://docs.scipy.org/doc/scipy/reference/generated/scipy.optimize.least_squares.html}},
  note         = {Accessed: 2026-05-01}
}

@inproceedings{blasch2019cyber,
  title={Cyber awareness trends in avionics},
  author={Blasch, Erik and Sabatini, Roberto and Roy, Aloke and Kramer, Kathleen A and Andrew, George and Schmidt, George T and Insaurralde, Carlos C and Fasano, Giancarmine},
  booktitle={2019 IEEE/AIAA 38th Digital Avionics Systems Conference (DASC)},
  pages={1--8},
  year={2019},
  organization={IEEE}
}

@inproceedings{tian2025gnss,
  title={GNSS Spoofing Identification with Receiver Motion-Induced Doppler (GSIR-MID)},
  author={Tian, Xin and Tian, Zijiao and Wang, Jiachen and Chen, Genshe and Pham, Khanh and Blasch, Erik},
  booktitle={Proceedings of the 38th International Technical Meeting of the Satellite Division of The Institute of Navigation (ION GNSS+ 2025)},
  pages={2085--2098},
  year={2025}
}

@article{shafique2021detecting,
  title={Detecting signal spoofing attack in UAVs using machine learning models},
  author={Shafique, Arslan and Mehmood, Abid and Elhadef, Mourad},
  journal={IEEE access},
  volume={9},
  pages={93803--93815},
  year={2021},
  publisher={IEEE}
}

\appendix

\section{IMM Filter Parameters}

The parameters used in the IMM filter and aircraft dynamic model are summarized in the following tables.

\begin{table}[h]
\centering
\caption{IMM Filter Parameters}
\label{tab:imm_parameters_appendix}
\begin{tabular}{lll}
\hline
\textbf{Parameter} & \textbf{Value} & \textbf{Description} \\
\hline
$\Delta t$ & $0.25$ s & Time step \\

$\bm{\mu}_0$ & $[0.45,\ 0.55]$ & Initial model probabilities \\

$\bm{\Pi}$ & 
$\begin{bmatrix}
0.98 & 0.02 \\
0.16 & 0.84
\end{bmatrix}$ 
& Model transition matrix \\

$\bm{Q}^{(\text{CV})}$ & 
\makecell{$\mathrm{diag}(2,\ 2,\ 2,$ \\ $120,\ 120,\ 120)$} 
& \makecell{CV process noise \\ (pos, vel)} \\

$\bm{Q}^{(\text{CA})}$ & 
\makecell{$\mathrm{diag}(4,\ 4,\ 4,$ \\ $50,\ 50,\ 50,$ \\ $20,\ 20,\ 20)$} 
& \makecell{CA process noise \\ (pos, vel, acc)} \\

$\bm{R}^{(\text{CV})}$ & 
$400\,\mathbf{I}_3$ 
& CV measurement noise \\

$\bm{R}^{(\text{CA})}$ & 
$50\,\mathbf{I}_3$ 
& CA measurement noise \\

\hline
\end{tabular}
\end{table}

\begin{table}[h]
\centering
\caption{Aircraft Model Parameters}
\label{tab:aircraft_parameters_appendix}
\begin{tabular}{lll}
\hline
\textbf{Parameter} & \textbf{Value} & \textbf{Description} \\
\hline
$I_{xx}$ & $0.2506$ kg$\cdot$m$^2$ & Moment of inertia about $x$-axis \\
$I_{yy}$ & $0.2506$ kg$\cdot$m$^2$ & Moment of inertia about $y$-axis \\
$I_{zz}$ & $1.0024$ kg$\cdot$m$^2$ & Moment of inertia about $z$-axis \\
$l$ & $0.635$ m & Arm length \\
$m_t$ & $12.66$ kg & Total mass of the aircraft \\
\hline
\end{tabular}
\end{table}

\end{document}